\documentclass[11pt,letterpaper]{article}
\usepackage{emnlp2017}
\usepackage{times}
\usepackage{latexsym}

\usepackage{subfigure} 

\usepackage{graphicx}
\usepackage{algorithm}
\usepackage{algorithmic}
\usepackage{amsmath}

\usepackage{url}

\emnlpfinalcopy

\def\emnlppaperid{***}

\title{End-to-End Multi-View    Networks for Text Classification}

\author{Hongyu Guo \and Colin Cherry \and Jiang Su \\
National Research Council Canada \\
1200 Montreal Road, Ottawa, Ontario, K1A 0R6, Canada \\
  {\tt firstname.lastname@nrc-cnrc.gc.ca}}

\date{}

\begin{document}

\maketitle

\begin{abstract}
We propose a multi-view  network for text classification. Our method automatically creates various views of its input text, each taking the form of soft attention weights that distribute the classifier's focus among a set of base features.
For a bag-of-words representation, each view focuses on a different subset of the text's words.
Aggregating many such views results in a more discriminative and robust representation. 
Through a novel architecture that both stacks and concatenates views, we produce a network that emphasizes both depth and width, allowing training to converge quickly.
Using our multi-view architecture, we establish new state-of-the-art accuracies on two benchmark tasks.
\end{abstract}

\section{Introduction}

State-of-the-art deep neural networks leverage task-specific architectures to develop hierarchical representations of their input, 
with each layer building a refined abstraction of the layer that came before it~\cite{DBLP:journals/corr/ConneauSBL16}.
For text classification, one can think of this as a single reader building up an increasingly refined understanding of the content.
In a departure from this philosophy, we propose a divide-and-conquer approach,
where a team of readers each focus on different aspects of the text, and then combine their representations to make a joint decision.

More precisely, the proposed Multi-View Network (MVN) for text classification learns to generate several views of its input text. Each view is formed by focusing on different sets of words through a view-specific attention mechanism. These views are arranged sequentially, so each subsequent view can
build upon or deviate from previous views as appropriate. The final representation that concatenates these diverse views should be more robust to noise than any one of its components. Furthermore, different sentences may look similar under one view but different under another, allowing the network to devote particular views to distinguishing between subtle differences in sentences, resulting in more discriminative representations.

Unlike existing multi-view neural network approaches for image processing~\cite{zhu2014multi,Su:2015:MCN:2919332.2919750}, where multiple views are provided as part of the input, our MVN learns to automatically create views from its input text by focusing on different sets of words. 
Compared to deep Convolutional Networks (CNN) for text~\cite{Zhang:2015:CCN:2969239.2969312,DBLP:journals/corr/ConneauSBL16}, the MVN strategy emphasizes network width over depth. Shorter connections between each view and the loss function enable better gradient flow in the networks, which makes the system easier to train. 
Our use of multiple views is similar in spirit to the weak learners used in  ensemble methods~\cite{Breiman:1996:BP:231986.231989,Friedman98additivelogistic,Wolpert92stackedgeneralization}, but our views produce vector-valued intermediate representations instead of classification scores,
and all our views are trained jointly with feedback from the final classifier.

\begin{figure}[tb]
	\centering
		\includegraphics[width=2.85in]{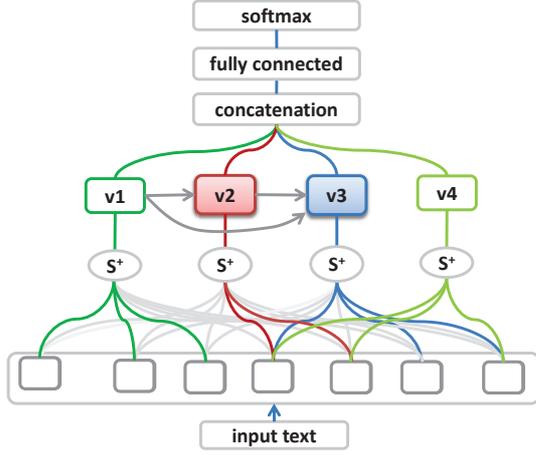}
	\label{fig:schema}
	\caption{A MVN architecture with four views.}
\end{figure}

Experiments on two benchmark data sets, the Stanford Sentiment Treebank~\cite{Socher2013} and the AG English news corpus~\cite{Zhang:2015:CCN:2969239.2969312}, show that 1) our method  achieves very competitive accuracy, 2) some views distinguish themselves from others by better categorizing specific classes, and 3) when our base bag-of-words feature set is augmented with convolutional features, the method establishes a new state-of-the-art for both data sets.


\section{Multi-View   Networks for Text} 
The MVN  architecture is depicted in Figure~\ref{fig:schema}. First, individual \textit{selection} vectors $s^{+}$ are created, each formed by a distinct softmax weighted sum over the word vectors of the input text. 
Next, these selections are sequentially transformed into \textit{views} $v$, with each view influencing the views that come after it.
Finally, all views are concatenated and fed into a two-layer perceptron for classification. 

\subsection{Multiple Attentions for Selection} \label{fea} Each selection $s^{+}$ is constructed by focusing on a different subset of  words from the original text, as determined by a softmax weighted sum~\cite{BahdanauCB14}.
Given a piece of text 
with $H$ words, we represent it as a bag-of-words feature matrix \mbox{$B$ $ \in {\rm I\!R}^{H \times d}$}. Each row of the matrix  corresponds to one word, which is represented by a  $d$-dimensional  vector, as provided by a learned word embedding table.
The  selection $s^{+}_{i}$ for the $i^{th}$ view is the softmax weighted sum of features:
\begin{align}
s^{+}_i =\sum^{H}_{h=1} d_{i,h}B[h:h]
\end{align}
where the weight $d_{i,h}$  is computed by:
\begin{align}
d_{i,h} =\frac{\exp (m_{i,h})}{\sum^{H}_{h=1} \exp (m_{i,h})} \end{align} \begin{align} m_{i,h} =w^{s}_i \tanh\left( W^s_i B[h:h]\right) \end{align} 
here, $w_i^s$ (a vector) and $W_i^s$ (a matrix) are learned selection parameters. By varying the weights $d_{i,h}$, the selection for each view can focus on different words from $B$, as illustrated by different color curves connecting to $s^{+}$ in Figure~\ref{fig:schema}. 

\subsection{Aggregating Selections into Views}

Having built one $s^{+}$ for each of our $V$ views, the actual views are then created as follows:
\begin{align}
v_{1}=&s^{+}_1 \mbox{ ; } v_{V}=s^{+}_V\\
v_{i} =&\tanh( W^v_{i}([v1;v2;...;v_{i-1};s^{+}_i])) \label{eq:viewrecur}\\ 
& \mbox{ for } i=2\ldots V-1 \nonumber
\end{align}
where $W^v_i$ are learned parameter matrices, and $[\ldots;\ldots]$ represents concatenation.
The first and last views are formed by solely $s^{+}$; however, they play very different roles in our network.
$v_{V}$ is completely disconnected from the others, an independent attempt at good feature selection, intended to increase view diversity~\cite{Muslea:2002:ASL:645531.655845,guo2006mining,guo2008multirelational,wang2015-deep-multi-view}.
Conversely, $v_1$ forms the base of a structure similar to a multi-layer perceptron with short-cutting, as defined by the recurrence in Equation~\ref{eq:viewrecur}.
Here, the concatenation of all previous views implements short-cutting, while the recursive definition of each view implements stacking, forming a deep network depicted by horizontal arrows in Figure~\ref{fig:schema}.
This structure makes each view aware of the information in those previous to it, allowing them to build upon each other.
Note that the $W^v$ matrices are view-specific and grow with each view, making the overall parameter count quadratic in the number of views.

\subsection{Classification with Views}

The final step is to transform our views into a classification of the input text. 
The MVN does so by concatenating its view vectors, which are then fed into a fully connected projection followed by a softmax function to produce a distribution over the  possible classes. 
Dropout regularization~\cite{abs-1207-0580} can be applied at this softmax layer, as in~\cite{Kim14}.

\subsection{Beyond Bags of Words}
\label{sec:CNNfeats}
The MVN's selection layer operates on a matrix of feature vectors $B$, which has thus far corresponded to a bag of word vectors.
Each view's selection makes intuitive sense when features correspond to words, as it is easy to imagine different readers of a text focusing on different words, with each reader arriving at a useful interpretation.
However, there is a wealth of knowledge on how to construct powerful feature representations for text, such as those used by convolutional neural networks (CNNs).
To demonstrate the utility of having views that weight arbitrary feature vectors, we  augment our bag-of-words representation with vectors built by $n$-gram filters max-pooled over the entire text~\cite{Kim14}, with one feature vector for each $n$-gram order, $n=2\ldots 5$.
The augmented $B$ matrix has $H+4$ rows.
Unlike our word vectors, the 4 CNN vectors each provide representations of the entire text.
Returning to our reader analogy, one could imagine these to correspond to quick ($n=2$) or careful ($n=5$) skims of the text.
Regardless of whether a feature vector is built by embedding table or by max-pooled $n$-gram filters, we always back-propagate through all feature construction layers, so they become specialized to our end task.

\section{Experiments}
\label{exp}
\subsection{Stanford Sentiment Treebank}
\label{setup}
The  Stanford  Sentiment  Treebank  contains  11,855 sentences from movie reviews.
We use the same splits for training, dev, and test data as in~\cite{Socher2013} to predict the fine-grained 5-class sentiment categories of the sentences.
For  comparison purposes, following~\cite{Kim14,DBLP:journals/corr/KalchbrennerGB14,DBLP:journals/corr/LeiBJ15}, we train  the models  using both phrases and sentences, but only evaluate sentences at test time.

We initialized all of the word embeddings~\cite{cherry2015unreasonable,cherry2015nrc} using the publicly available 300 dimensional pre-trained  vectors from GloVe~\cite{pennington2014glove}. We learned 8 views with 200 dimensions each, which requires us to project the 300 dimensional word vectors, which we implemented using a linear transformation, whose weight matrix and bias term are shared across all words, followed by a $\tanh$ activation. For optimization, we used Adadelta~\cite{DBLP:journals/corr/abs-1212-5701}, with a starting learning
rate of 0.0005 and a mini-batch of size 50. Also, we used dropout (with a rate of 0.2) to avoid overfitting. 
All of these MVN hyperparameters were determined through experiments measuring validation-set accuracy.

\begin{table}[t]
  \centering
\begin{tabular}{l|c}\hline
MVN (with convolutional features) & \textbf{51.5}\\
MVN &49.6 \\\hline
high-order CNN& 51.2\\
tree-LSTM& 51.0\\ 
DRNN& 49.8 \\
DCNN& 48.5 \\
CNN-MC& 47.4 \\
NBoW&  44.5 \\
SVM & 38.3 \\
 \hline
\end{tabular}
  \caption{Accuracies on the Stanford Sentiment Treebank 5-class classification task; except for the MVN, all results are drawn from~\cite{DBLP:journals/corr/LeiBJ15}.}
  \label{tab:accuracy:sen}
\end{table}

The test-set accuracies obtained by different learning methods, including the current state-of-the-art results, are presented in Table~\ref{tab:accuracy:sen}. 
The results indicate that the bag-of-words MVN outperforms most methods, but obtains lower accuracy than the state-of-the-art results achieved by the tree-LSTM~\cite{DBLP:journals/corr/TaiSM15,zhu2015long} and the high-order CNN~\cite{DBLP:journals/corr/LeiBJ15}. However, when augmented with 4 convolutional features as described in Section~\ref{sec:CNNfeats}, the MVN strategy surpasses both of these, establishing a new state-of-the-art on this benchmark.

\begin{figure}
	\centering
		\includegraphics[width=2.843in]{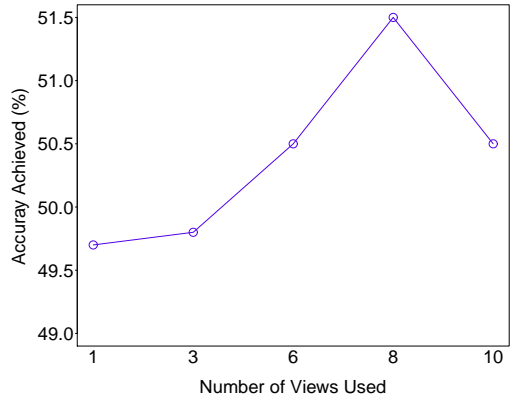}
      \caption{Accuracies obtained by varying the number of views.}
      \label{fig:view}
\end{figure}

In Figure~\ref{fig:view}, we present the test-set accuracies obtained while varying the number of views in our MVN with convolutional features. These results indicate that better predictive accuracy can be achieved while increasing the number of views up to eight. 
After eight, the accuracy starts to drop. 
The number of MVN views should be tuned for each new application, but it is good to see that not too many views are required to achieve optimal performance on this task.

\begin{figure}
	\centering
		\includegraphics[width=2.843in]{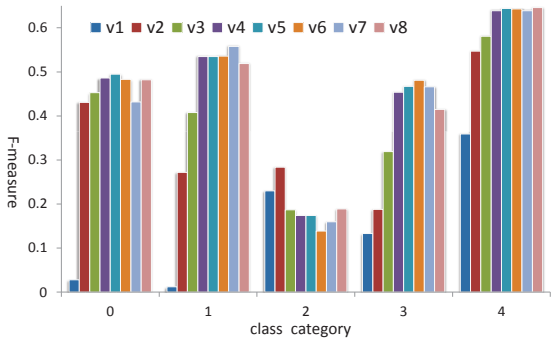}
      \caption{Class-specific F-measures obtained by Na{\"i}ve Bayes classifiers built over different views.}
      \label{fig:fmeasure}
\end{figure}

To better understand the benefits of the MVN method, we further analyzed the eight views constructed by our best model. After training, we obtained the view representation vectors for both the training and testing data, and then independently trained a very simple, but fast  and stable Na{\"i}ve Bayes classifier~\cite{McCallum98acomparison} for each view.  We report class-specific F-measures for each view in Figure~\ref{fig:fmeasure}. From this figure, we can observe that different views focus on different target classes. For example, the first two views perform poorly on the 0 (very negative) and 1 (negative) classes, but achieve the highest F-measures on the 2 (neutral) class. 
Meanwhile, 
the non-neutral classes each have a different view that achieves the highest F-measure.
This suggests that some views have specialized in order to better separate subsets of the training data.


\begin{table}[t]
  \centering
\begin{tabular}{l|c}\hline
Full MVN & 51.5\\\hline
Voting by 8 independent, 1-view MVNs & 50.2 \\
(weak-learner: 49.5 $\pm$ 0.20)\\
MVN w/ no horizontal links & 49.0\\
MVN w/ length-1 horizontal links & 50.5\\
 \hline
\end{tabular}
  \caption{Ablation experiments on the Stanford Sentiment Treebank test set}
  \label{tab:ablation}
\end{table}

We provide an ablation study in Table~\ref{tab:ablation}.
First, we construct a traditional ensemble model.
We independently train eight MVN models, each with a single view, to serve as weak learners.
We have them vote with equal weight for the final classification, obtaining a  test-set accuracy of 50.2.    
Next, we restrict the views in the MVN to be unaware of each other. That is, we  replace Equation~\ref{eq:viewrecur} with $v_i=s^{+}_i$, which removes all horizontal links in Figure~\ref{fig:schema}.
This drops performance to 49.0.
Finally, we  experiment with a variant of  MVN, where  each view is only connected to the most recent previous view,
replacing Equation~\ref{eq:viewrecur} with $v_{i} =\tanh( W^v_{i}([v_{i-1};s^{+}_i]))$,
leading to a version where the parameter count grows linearly in the number of views.
This drops the test-set performance to 50.5. 
These  experiments suggest that  enabling the views to  build upon each other is crucial for achieving the best performance.

\subsection{AG's English News Categorization}

\begin{table}[t]
  \centering
\begin{tabular}{l|c}\hline
MVN (with convolutional features)& \textbf{7.13}\\
MVN & 7.49\\\hline 
 $n$-grams TFIDF &7.64 \\
 $n$-grams &7.96 \\
Lg. Lk. Convolution &8.55 \\
29 layers Convolution with KMaxPooling &8.67 \\
Lg. Full Convolution &9.85\\
BoW &11.19 \\
LSTM &13.94 \\
Bag-of-means &16.91 \\
 \hline
\end{tabular}
  \caption{Error rates on the AG News test set. All results except for the MVN are drawn from~\cite{DBLP:journals/corr/ConneauSBL16}}
  \label{tab:accuracy2}
\end{table}


The AG corpus~\cite{Zhang:2015:CCN:2969239.2969312,DBLP:journals/corr/ConneauSBL16} contains categorized news articles 
 from more than 2,000 news outlets on the web. The task has four classes, and for each class there are 30,000 training documents and 1,900 test documents. A random sample of the training set was used for hyper-parameter tuning.
The training and testing settings of this task are exactly the same as those presented for the Stanford Sentiment Treebank task in Section~\ref{setup}, except that the mini-batch size is reduced to 23, and each view has a dimension of 100.

The test errors obtained by various methods are presented in Table~\ref{tab:accuracy2}. These results show that the bag-of-words MVN outperforms the state-of-the-art accuracy obtained by the non-neural $n$-gram TFIDF approach~\cite{Zhang:2015:CCN:2969239.2969312}, as well as several very deep CNNs~\cite{DBLP:journals/corr/ConneauSBL16}. Accuracy was further improved when the MVN was augmented with 4 convolutional features.

\begin{figure}[t]
	\caption{Accuracies and cost on the validation set during training on the AG News data set.}
	\centering
		\includegraphics[width=3in]{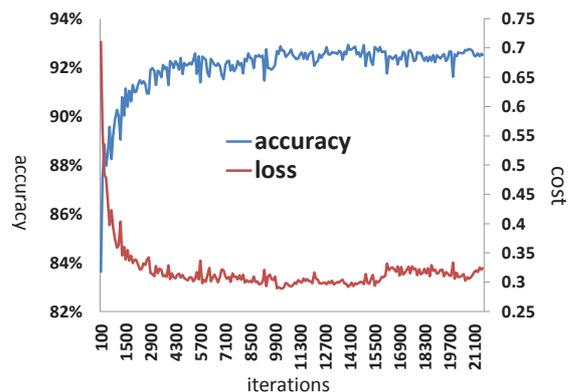}
	\label{fig:trainingcurve}
\end{figure}

In Figure~\ref{fig:trainingcurve}, we show how accuracy and loss evolve on the validation set during MVN training. These curves show that training is quite stable. The MVN achieves its best results in just a few thousand iterations.

\section{Conclusion and Future Work}
We have presented a novel multi-view neural network for text classification, which creates multiple views of the input text, each represented as a weighted sum of a base set of feature vectors.
These views work together to produce a discriminative feature representation for text classification.  
Unlike many neural approaches to classification, our architecture emphasizes network width in addition to depth, enhancing gradient flow during training.
We have used the multi-view network architecture to establish new state-of-the-art results on two benchmark text classification tasks.
In the future, we wish to better understand the benefits of generating multiple views, explore new sources of base features, and apply this technique to other NLP problems such as translation or tagging.

\bibliography{reference}

\bibliographystyle{emnlp_natbib}

\end{document}